# Personalization of Affective Models to Enable Neuropsychiatric Digital Precision Health Interventions: A Feasibility Study

Ali Kargarandehkordi

*Information and Computer Sciences Department*
*Honolulu, Hawai'i, USA*
*Email: kargaran@hawaii.edu*

Matti Kaisti

*Department of Computing, University of Turku,*
*Turku, Finland*
*Email: mkaist@utu.fi*

*Peter Washington

*Information and Computer Sciences Department*
*Honolulu, Hawai'i, USA*
*Email: pyw@hawaii.edu*

Mobile digital therapeutics for autism spectrum disorder (ASD) often target emotion recognition and evocation, which is a challenge for children with ASD. While such mobile applications often use computer vision machine learning (ML) models to guide the adaptive nature of the digital intervention, a single model is usually deployed and applied to all children. Here, we explore the potential of model personalization, or training a single emotion recognition model per person, to improve the performance of these underlying emotion recognition models used to guide digital health therapies for children with ASD. We conducted experiments on the Emognition dataset, a video dataset of human subjects evoking a series of emotions. For a subset of 10 individuals in the dataset with a sufficient representation of at least two ground truth emotion labels, we trained a personalized version of three classical ML models (k-nearest neighbors, random forests and a dense neural network) on a set of 51 features extracted from each video frame. We ensured that all frames used to train the models occurred earlier in the video than the frames used to test the model. We measured the importance of each facial feature for all personalized models and observed differing ranked lists of top features across subjects, motivating the need for model personalization. We then compared the personalized models against a generalized model trained using data from all 10 participants. The mean F1-scores achieved by the personalized models were 90.48%, 92.66%, and 86.40%, respectively. By contrast, the mean F1-scores reached by non-personalized models trained on different human subjects and evaluated using the same test set were 88.55%, 91.78%, and 80.42%, respectively. The personalized models outperformed the generalized models for 7 out of 10 participants. PCA analyses on the remaining 3 participants revealed relatively little facial configuration differences between emotion labels within each subhect, suggesting that personalized ML will fail when the variation among data points within a subject's data is too low. This preliminary feasibility study demonstrates the potential as well as the ongoing challenges with implementing personalized models which predict highly subjective outcomes, an increasingly common task in digital health applications within psychiatry and the behavioral sciences.

*Keywords:* personalized machine learning, explainable AI, affective computing, digital health, digital therapeutics, neuropsychiatric digital therapies, autism spectrum disorder, precision health, digital phenotyping

---

* Corresponding author

## 1. Introduction

Emotions may be conveyed through a combination of facial expressions, vocalizations, gestures, and body movements [1-3]. Both the evocation and the understanding of emotional expressions play a crucial role in detecting certain types of developmental disorders. Autism spectrum disorder (ASD) affects almost 1 in 44 people in America [4], and it is the fastest-growing developmental disorder in the United States [5-6]. ASD is a multifaceted neuropsychiatric disorder that appears in diverse phenotypic forms. Children with autism tend to evoke emotions differently than neurotypical peers, and they find it challenging to identify facial expressions conveyed by other individuals [7-9].

To improve social communication of children with ASD, a variety of AI-powered mobile digital therapeutics have been developed which target emotion expression in particular [10-18]. These digital health innovations consist of smartphone apps and wearable devices that enable families to provide therapy in the comfort of their home setting with the ability to customize the intervention structure to suit their child's needs [19-25]. For example, Superpower Glass [19-20, 22-26] is an artificial intelligence (AI) powered digital therapeutic designed to aid children in understanding emotion evocations by conversation partners by providing real time feedback from a facial expression recognition model. The therapeutic operates on a Google Glass connected to a smartphone and provides real-time social cues to children with ASD. "Guess What" [18] is another digital therapy encouraging, among other therapeutic behaviors, increased emotion expression using a Charades-style mobile game. Although considerable progress has been made in providing sensitive and specific emotion expression feedback to children using such digital health therapeutics, there remain several technical challenges that must be addressed to facilitate near-perfect performance. Many existing works in this area rely on a one-size-fits-all emotion recognition computer vision model. However, this approach may overlook individual variations in emotional expressions and could result in less accurate assessments for certain individuals. Additionally, factors such as cultural disparities, age, and personal characteristics can influence emotional expression, posing further challenges to the effectiveness of the generic model.

Personalized models, or the creation of a separate AI model per person, offer the advantage of tailoring the emotional assessment and therapy process to each individual's unique facial dynamics. By primarily considering the specific emotional nuances of the individual, these personalized models enhance the accuracy of emotion recognition, resulting in more effective outcomes for digital therapeutics and digital phenotyping. Academic evidence supports the idea that personalization can enhance the performance of emotion recognition systems [27-29]. Personalized machine learning (ML) techniques have the potential to unlock more precise and context-aware emotion recognition capabilities compared to the traditional paradigm of using generic models.

Here, we study personalization on emotion recognition models using a video dataset called Emognition. As a stride towards making our models explainable, we focus in this paper on the feature extraction of interpretable facial features fed into classical ML models rather than using convolutional neural networks to automatically learn complex features. We train separate models per human subject and evaluate on that individual's data, ensuring that all data in the training set occurs earlier temporally than the evaluation set. We find that the personalized models outperform baseline models trained on baseline models using data from other subjects in the dataset, indicating that model personalization can benefit the performance of precision digital health that involve complex and potentially subjective automated quantifications of the end user.

## 2. Methods

We created personalized emotion recognition models (Figure 1) for a targeted subset of human subjects in the Emognition dataset (details below). We trained three models, namely k-nearest neighbors, random forest, and a dense neural network, using 51 extracted interpretable facial features from each video frame. A key focus of our experimental setup is to compare the performance of personalized vs. non-personalized (generic) models.

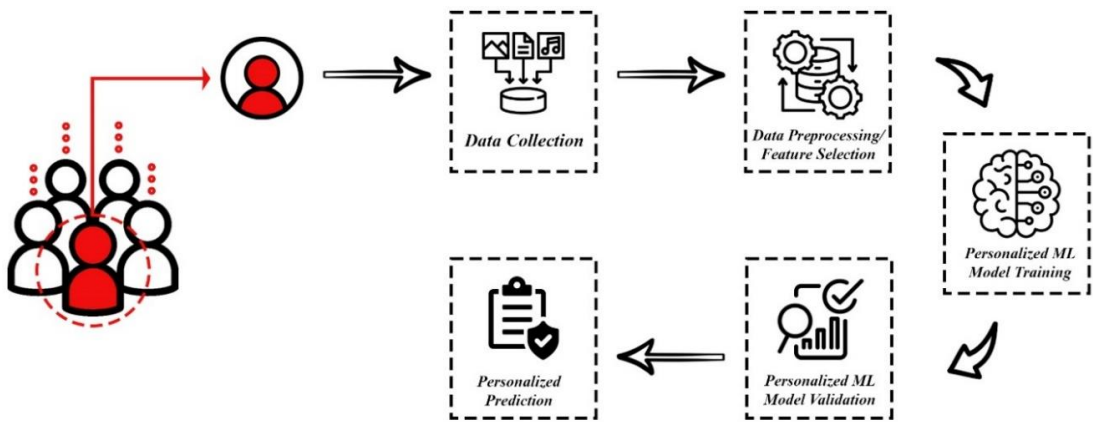

Fig. 1. Personalized ML workflow. Rather than training a traditional one-size-fits-all model, we propose the development of a single model per individual. While we evaluate this procedure for affective computing, this paradigm can be applied to precision health more broadly.

### 2.1. *Emognition Dataset*

The Emognition dataset [30] encompasses data from 43 participants aged between 19 and 29, including 21 females, who were exposed to emotionally stimulating film clips specifically designed to evoke ten distinct emotions (Table 1). Facial features were automatically extracted using the OpenFace toolkit [31] (version 2.2.0, default parameters) and Quantum Sense software (Research Edition 2017, Quantum CX, Poland). The OpenFace library provides essential facial landmark points and action units' values while the Quantum Sense software identifies fundamental emotions including neutral, anger, disgust, happiness, sadness, surprise (Table 1) and head pose.

The emotion label population and distribution varied between subjects. The dataset for some participants contained insufficient ground truth emotion labels even for a fair binary classification, possibly due to challenges in recognizing emotions during the video stimulus because of mostly maintaining a neutral state by the participant. To ensure a more equitable distribution of labels for the classification task, we selected a subset of 10 participants who had a sufficient number of labels for at least 2 emotions. Because the >=2 emotions available in the dataset varied across these 10 participants, we performed different classification tasks based on the label population for each emotion, ranging from binary classification to the classification of all six emotion labels.

## 2.2. Data Preprocessing and Arrangement

We observed varying elicited emotional expressions across participants. For example, some individuals had no labels for certain emotion stimulus videos (e.g., no anger labels for anger video, no surprise label for surprise video, etc.), indicating insufficient facial expression stimulation. This disparity in label counts suggests that the threshold and manner of eliciting each emotion differ between individuals. The distribution of labels for each emotion across 10 stimulus videos and a neutral video for one demonstrative participant (no. 22) is shown in Table 1.

Table 1. The discrepancy between the emotions evoked (Actual) for one demonstrative participant vs. the emotional stimulus provided (Prompt). We use the emotions evoked for our ground truth labels.

| Prompt / Actual | Amusement | Anger | Awe | Disgust | Enthusiasm | Fear | Liking | Neutral | Sadness | Surprise |
|---|---|---|---|---|---|---|---|---|---|---|
| Anger | 0 | 0 | 0 | 0 | 0 | 3 | 0 | 0 | 0 | 0 |
| Disgust | 0 | 0 | 0 | 0 | 0 | 0 | 0 | 0 | 0 | 0 |
| Sadness | 398 | 213 | 135 | 227 | 87 | 111 | 128 | 1812 | 9 | 159 |
| Neutral | 6766 | 6752 | 6822 | 3339 | 7036 | 6993 | 6541 | 5468 | 7126 | 966 |
| Surprise | 0 | 37 | 0 | 0 | 0 | 116 | 1 | 0 | 0 | 0 |
| Happiness | 40 | 198 | 0 | 495 | 29 | 0 | 6 | 0 | 0 | 1832 |

Notably, there were no labels for anger, surprise, and disgust emotions in participant 22's respective video stimulus experiments. Additionally, the dataset contained only a few labels for anger and no labels for disgust at all, presenting challenges in achieving a balanced dataset for training and classification. To address this challenge, we created separate models for specific recognition tasks. To overcome the label imbalance, we collected all emotion labels specific to each discrete emotion from several video stimulus experiments into a single dataset for training each subject's model. To mitigate the impact of imbalanced label distribution, we carefully curated emotions with a substantial number of labels for accurate and unbiased classification results. For example, for participant no. 22, we trained a model to only recognize Sadness, Neutral, and Happiness, which were the only well-represented labels for this participant. To ensure a robust analysis, we sampled a reduced balanced subset of 1600 instances for each of the emotions for ML model training.

We also trained a series of one-size-fits-all models to establish baselines for comparison with the personalized models. We selected and consolidated data from all 10 participants to train the one-size-fits-all models. We trained a separate model for each combination of emotions sufficiently represented by a participant (e.g., a one-size-fits-all model for happy vs. sad vs. neutral, a one-size-fits-all model for happy vs. neutral, etc).

To provide further clarity, throughout the paper, we will refer to the resulting datasets from these procedures as the "personalized dataset" and the "generic dataset".

### 2.3. *Feature Extraction and Selection*

As a stride towards the development of interpretable models, we identified eye gaze, eye landmarks, pose landmarks, and Action Unit (AU) features which displayed the most predictive saliency. Eye gaze and landmarks consist of x, y, and z components. We aggregated features by calculating the mean values of each feature within each positional direction, resulting in a single combined feature value. To normalize the data, we implemented z-score standardization.

We explored the relationships between each feature and the target variable. The scatterplots for all feature combinations revealed intricate patterns, correlations, and trends, providing valuable insights for feature selection and modeling. A pairwise feature correlation matrix for a demonstrative user (Figure 2) uncovered complex data relationships, including noticeable linear and nonlinear associations between various pairs of variables, such as the average y-axis coordination values of pose and eye landmarks.

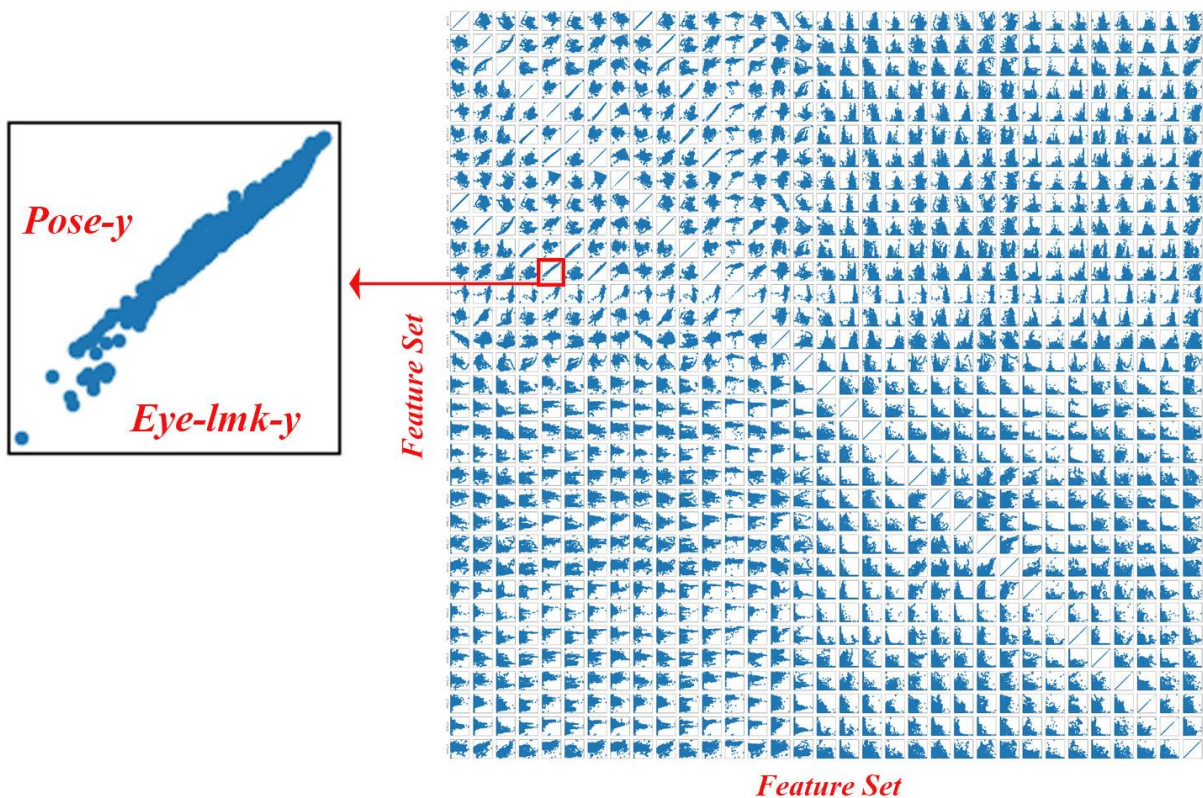

Fig. 2. Pairwise feature correlation matrix for a demonstrative participant. We plot each feature against value against every other feature to observe correlations between features.

We applied Principal Component Analysis (PCA) to the features and color-coded each data point (i.e., video frame) based on the corresponding emotion (Figure 3 displays a demonstrative example for 2 participants. When comparing the plots for both the personalized dataset (one individual) and generic dataset (all ten individuals), we observed a clear difference in the separation of data points relevant to each emotion class. We observe that the personalized dataset consistently yielded a better cluster visualization compared to the generic model, suggesting the potential of the downstream ML models to result in superior performance.

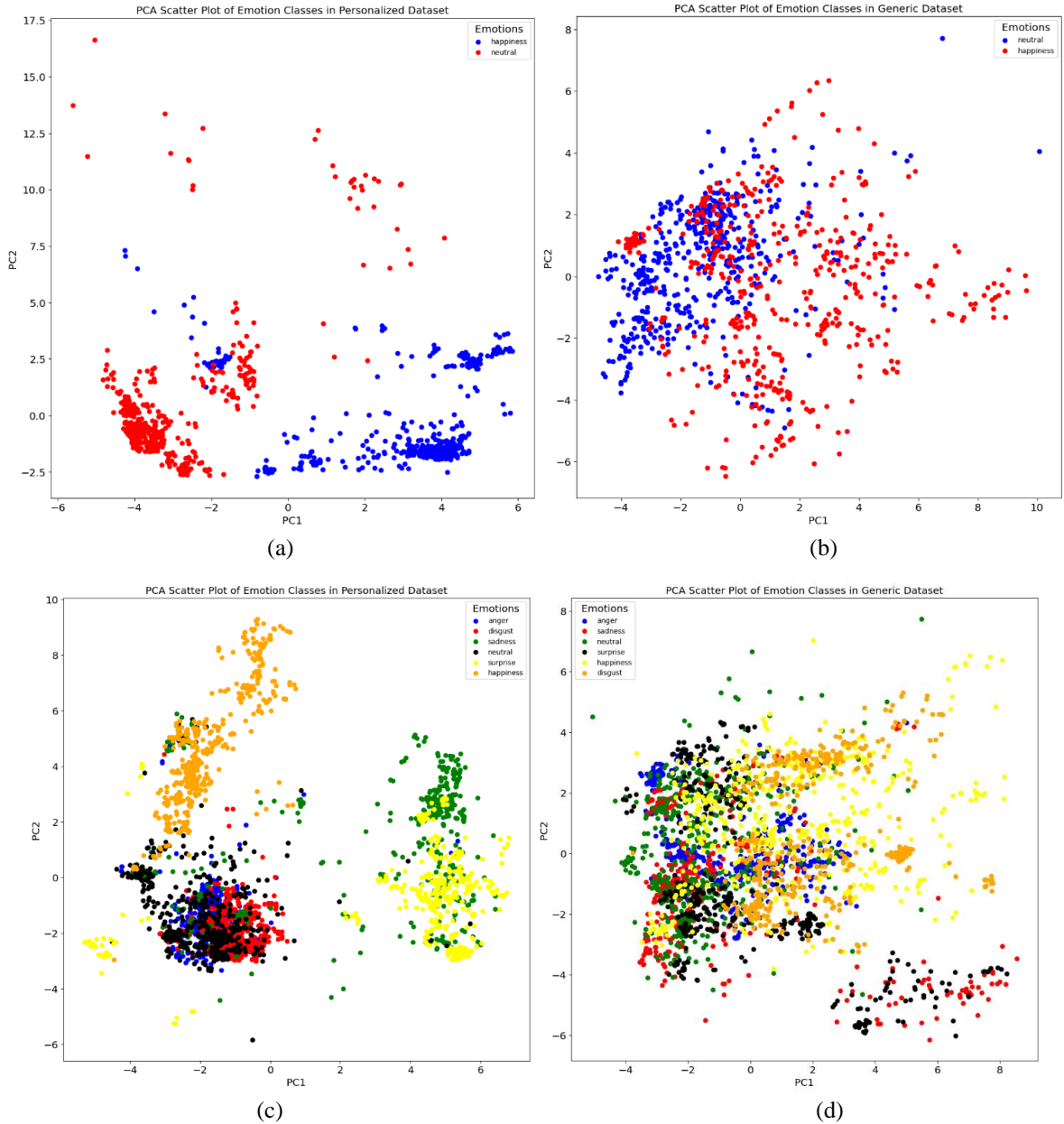

Fig. 3. PCA visualizations of the personalized dataset for two participants (a), (c) as well as the corresponding generalized dataset (b), (d).

### 2.4. *Model Selection*

We trained personalized and generic versions of three classical ML models for all 10 participants: K-Nearest Neighbors (KNN), Random Forest, and Multilayer Perceptron (MLP).

We conducted cross-validated grid search for all 3 models to systematically optimize hyperparameters.

## 2.5. *Evaluation*

We performed a nested cross-validation procedure to simultaneously optimize hyperparameters and assess each classifier's performance. We performed hyperparameter tuning using grid search with an inner cross-validation of 5 folds. The best model for each classifier was selected based on evaluating different set of hyperparameters, and its performance was evaluated on the test data using 10-fold outer cross-validation. This process ensured rigorous optimization of the classifier and comprehensive assessment of its classification performance. We used both AUC - ROC and F1-score as our primary evaluation metrics in a one-vs-rest multiclass approach.

## 3. Results

We conducted a comprehensive evaluation of the models, including the KNN classifier, Random Forest, and MLP classifier, employing a rigorous procedure involving component setup, cross-validation, and performance assessment. In the personalized experiment, the KNN model achieved an average F1-score of 0.904, while in the generic experiment, it attained an average F1-score of 0.885. Similarly, the Random Forest yielded average F1-scores of 0.926 and 0.917 in the personalized and generic experiments, respectively, while the MLP classifier obtained average F1-scores of 0.864 and 0.804 for the personalized and generic experiments, respectively.

We directly compare the F1-score of the personalized vs. generic models for all 10 participants (Table 2). The personalized ML approach outperformed the general-purpose models for emotion recognition in 7 out of 10 individuals.

Table 2. Overall performance (F1-score) of the models in both personalized and generic approaches on the same evaluation task. Participants whose performance on the personalized model was lower than for the generic model are highlighted in red.

| Participant | F1-score (Personalized) | | | F1-score (Generic) | | |
|---|---|---|---|---|---|---|
| | KNN | RF | DNN | KNN | RF | DNN |
| **No. 22** | 93.1% | 95.3% | 88.2% | 86.9% | 91.4% | 76.7% |
| **No. 25** | 89.3% | 91.5% | 83.4% | 88.0% | 92.0% | 81.2% |
| **No. 28** | 96.6% | 97.8% | 93.7% | 88.5% | 92.7% | 82.9% |
| **No. 29** | 83.6% | 86.3% | 78.6% | 90.0% | 92.0% | 84.4% |
| **No. 32** | 93.4% | 95.1% | 91.8% | 92.6% | 95.0% | 88.0% |
| **No. 39** | 87.2% | 90.0% | 83.8% | 92.1% | 94.0% | 87.0% |
| **No. 40** | 99.6% | 99.9% | 97.2% | 88.3% | 91.0% | 78.4% |
| **No. 42** | 93.2% | 94.7% | 89.9% | 87.2% | 91.6% | 78.0% |
| **No. 45** | 82.5% | 86.3% | 73.3% | 82.2% | 87.1% | 63.9% |
| **No. 48** | 86.3% | 89.7% | 84.1% | 89.7% | 91.0% | 83.7% |

Analyzing the ROC curves and confusion matrices for a demonstrative participant (no. 22) provides insights into the personalized models' classification capabilities (Figure 5). Retaining the best hyperparameter combinations aided in identifying optimal settings and understanding model behavior.

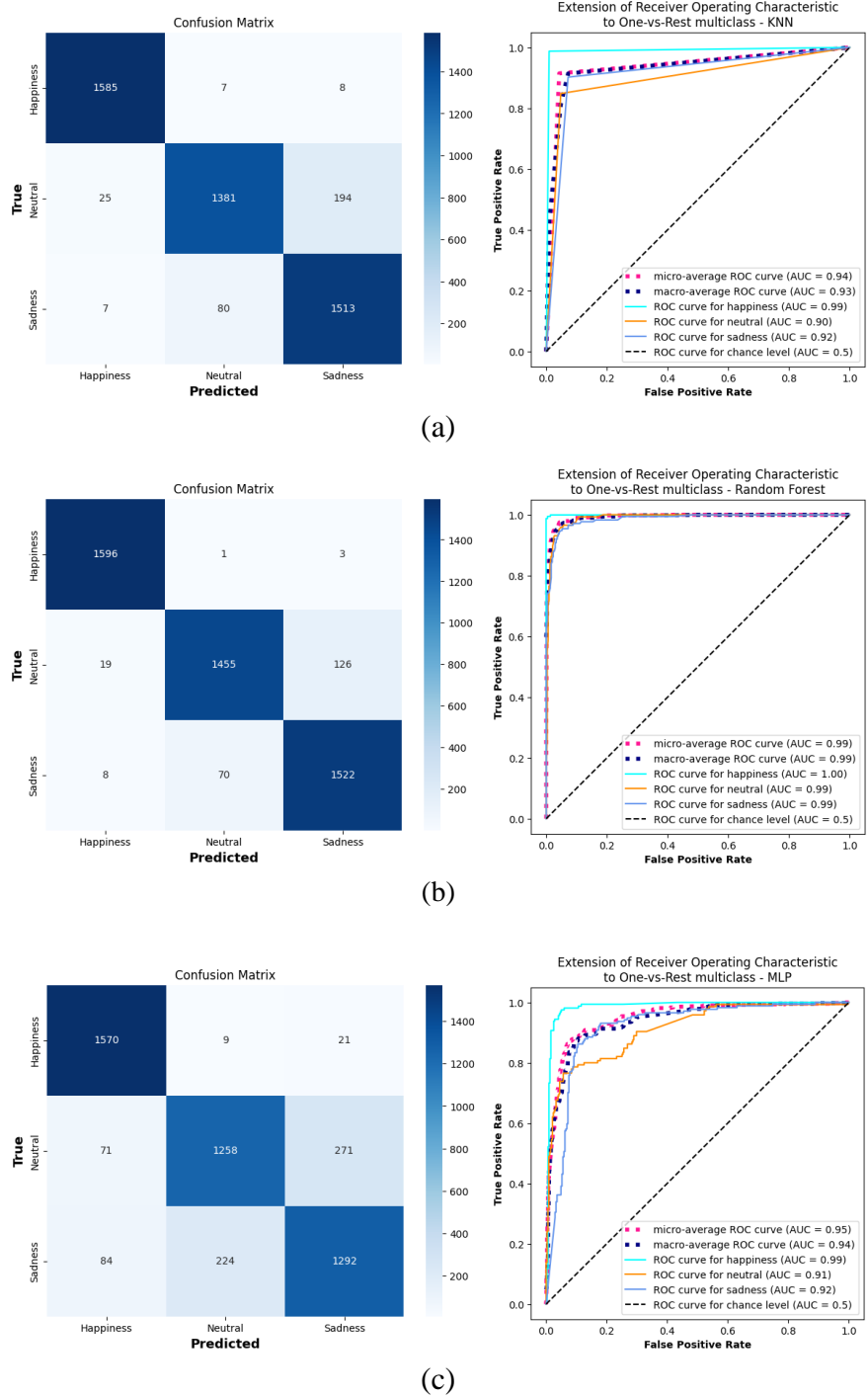

Fig. 5. Confusion matrices and ROC curves for 3 separate models, (a) KNN, (b) random forest, and (c) dense neural network, each trained and evaluated on a demonstrative participant, no. 22.

In certain instances, the performance of personalized models did not surpass that of the generic models (Table 2). To investigate this discrepancy, we analyzed the PCA plots for these individuals. The PCA plots revealed that the data points representing emotional states did not form distinct clusters for these participants, especially with respect to the separation of the generalized dataset containing data from all 10 participants (e.g., Figure 6). Consequently, it is foreseeable that personalized models might encounter difficulty in accurately discerning individual emotional states compared to generic models in cases where the individual makes relatively little variation in their facial movement across emotions.

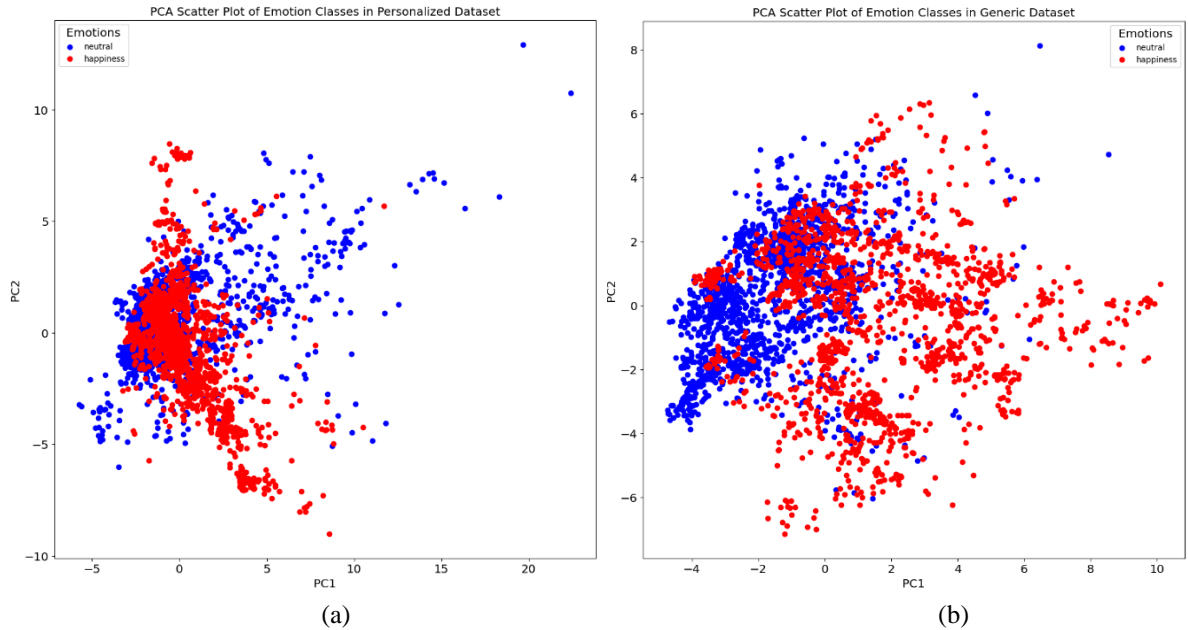

Fig. 6. PCA for participant no. 39. The lack of separability in the personalize dataset (a) compared to the generalized dataset for the happy vs. neutral data (b) provides an explanation for the lack of performance gain in this instance.

To explore of the most salient features contributing to precise emotion classification, we computed the impurity-based importance of each feature of the Random Forest model. This feature ranking approach inherently accounts for the correlations and complex nonlinear relationships between features. We plot the impurity-based importance of each feature in Figure 4 for a demonstrative set of 3 users whose ML models were all trained to predict Happy vs. Neutral. We observe that the top-ranked features across participants vastly differ, further supporting the need for model personalization in affective computing.

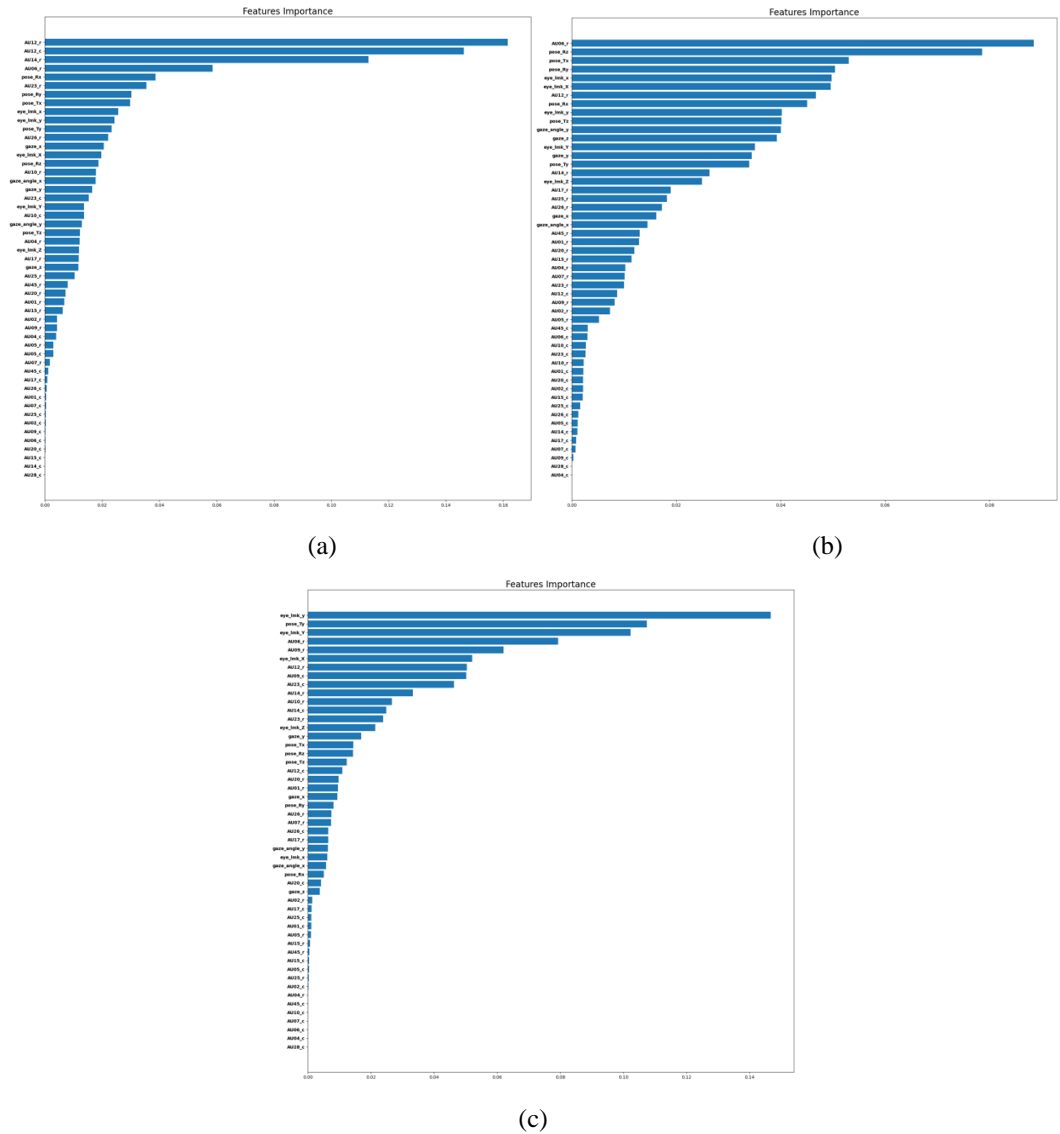

Fig. 4. Impurity-based importance of each facial feature according to the random forest model for 3 participants' personalized models. The model for all 3 participants is predicting Happy vs. Neutral.

## 4. Discussion

In most cases, the personalized ML approach demonstrated a slightly stronger ability to distinguish the nuances of each participant's emotion expressions compared to the general-purpose models. In cases where an improved performance was not observed for a participant, PCA revealed a lack of sufficient data separation for the participant's expressions with respect to the general-purpose models. These results support the effectiveness of the personalized ML approach in classifying emotions and highlight its potential for further advancements in the field of emotion recognition. While we evaluated model personalization on the affective computing task, which has direct implications for digital health therapies for children with ASD, this paradigm can be applied to a variety of precision health tasks where a property of interest (e.g., blood glucose levels in diabetes patients) should be predicted repeatedly for a single end user.

Although convolutional neural networks and vision transformers offer the possibility of better performance gains, we deliberately opted to use classical ML methods to prioritize the interpretability of our machine learning models. While these state-of-the-art models have demonstrated remarkable success in image recognition, their complex architecture often renders them as "blackbox" models, making it challenging to interpret and understand the learned features influencing their predictions. By contrast, the automatic feature extraction we performed enabled us to inspect and comprehend the specific facial features that contribute to emotion recognition between subjects. Notably, we learned that the top facial features in the personalized models differed across subjects, highlighting the need for personalized ML.

Our study, while demonstrating promise for personalized learning of relatively subjective tasks like affective computing, contains several limitations and can therefore only be considered as a feasibility study. We evaluated our method on only 10 participants. While this experimental paradigm can be viewed as 10 independent N=1 studies, we hope to expand this set of experiments in future work to more and larger datasets.

An especially promising avenue of future work is the exploration of self-supervised pre-training to enhance the personalization capabilities of deep learning models. By pre-training deep learning models on large and diverse datasets using self-supervised learning, each personalized model can learn the baseline dynamics of each individual's face without any training labels. These pre-trained models can then be fine-tuned with relatively few labeled examples. We note that this self-supervised learning paradigm would only be possible with a deep learning model rather than the classical ML approaches we present. There is a clear tradeoff between interpretability and performance.